\documentclass[10pt,twocolumn,letterpaper]{article}

\usepackage[pagenumbers]{cvpr} 

\usepackage{siunitx}
\usepackage{makecell}
\usepackage{multirow}
\usepackage{colortbl}
\usepackage{indentfirst}
\usepackage[ruled,linesnumbered]{algorithm2e}
\usepackage{stfloats}

\definecolor{cvprblue}{rgb}{0.21,0.49,0.74}
\usepackage[pagebackref,breaklinks,colorlinks,allcolors=cvprblue]{hyperref}

\def\paperID{5880} 
\def\confName{CVPR}
\def\confYear{2025}

\title{Learning on Less: Constraining Pre-trained Model Learning for Generalizable Diffusion-Generated Image Detection}

\author{
	Yingjian Chen$^{1}$, Lei Zhang$^{1}$, Yakun Niu$^{1 }$\thanks{Corresponding author}, Lei Tan$^{1}$, Pei Chen$^{1}$\\
	$^{1}$Henan Key Laboratory of Big Data Analysis and Processing, Henan University\\
	{\tt\small \{yingjianchen, zhanglei, ykniu, chenpei, tanlei\}@henu.edu.cn}
}

\begin{document}
\maketitle
\begin{abstract}	
Diffusion Models enable realistic image generation, raising the risk of misinformation and eroding public trust. Currently, detecting images generated by unseen diffusion models remains challenging due to the limited generalization capabilities of existing methods. To address this issue, we rethink the effectiveness of pre-trained models trained on large-scale, real-world images. Our findings indicate that: 1) Pre-trained models can cluster the features of real images effectively. 2) Models with pre-trained weights can approximate an optimal generalization solution at a specific training step, but it is extremely unstable. Based on these facts, we propose a simple yet effective training method called Learning on Less (\textbf{LoL}). LoL utilizes a random masking mechanism to constrain the model’s learning of the unique patterns specific to a certain type of diffusion model, allowing it to focus on less image content. This leverages the inherent strengths of pre-trained weights while enabling a more stable approach to optimal generalization, which results in the extraction of a universal feature that differentiates various diffusion-generated images from real images. Extensive experiments on the GenImage benchmark demonstrate the remarkable generalization capability of our proposed LoL. With just \textbf{1\%} training data, LoL significantly outperforms the current state-of-the-art, achieving a \textbf{13.6\%} improvement in average ACC across images generated by eight different models.
\end{abstract}
    
\section{Introduction}
\label{sec:intro}

\begin{figure}[t]
	\centering
	\includegraphics[width=8.2cm]{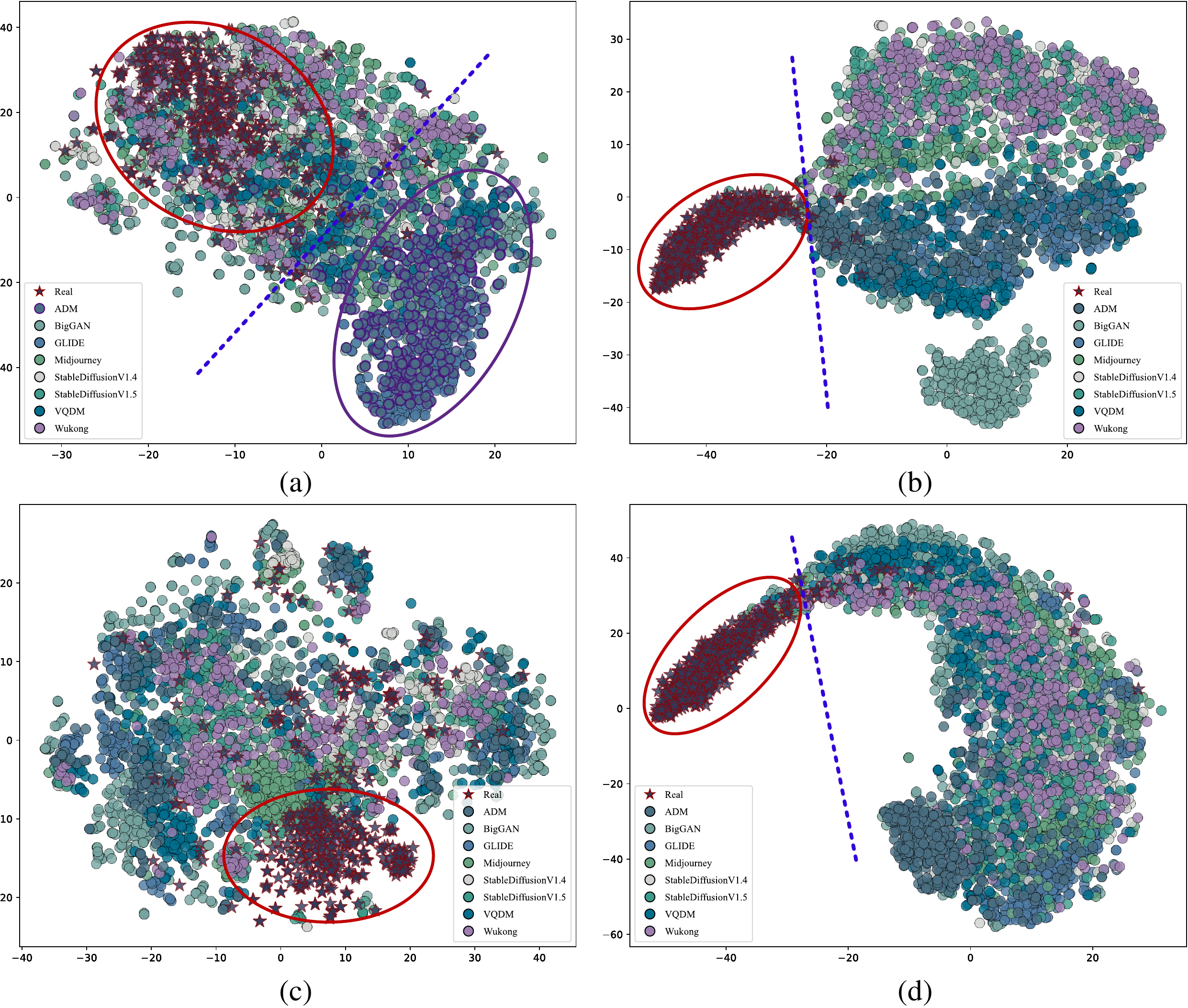}
	\caption{t-SNE visualization \cite{van2008visualizing} of features from images generated by eight models in the GenImage dataset.
		(a) and (d) Feature space of a standard ResNet50 and our proposed method trained on images generated by a single diffusion model type (ADM).
		(b) Feature space of a standard ResNet50 trained on images generated by all involved model types.
		(c) Feature space of a zero-shot pre-trained CLIP-ResNet50. }
	\label{img3}
\end{figure}

In recent years, the rapid development of diffusion models has significantly enhanced image generation quality. Approaches like DDPM (Denoising Diffusion Probabilistic Models) \cite{ho2020denoising} and DDIM (Denoising Diffusion Implicit Models) \cite{song2020denoising} can easily produce highly realistic and high-quality images. Specially, in the domain of Text-to-Image generation, diffusion models \cite{ramesh2022hierarchical, kim2022diffusionclip, saharia2022photorealistic} integrate the diffusion process with advanced text embedding models to generate semantically consistent and realistic images from textual descriptions.  Methods such as \cite{nichol2021glide, rombach2022high} have further refined the image generation process, enabling a more accurate representation of detailed information from the text in the generated images. 
However, the rapid development of diffusion models has made it increasingly easy to generate fake images. For instance, online platforms such as MidJourney \cite{midjourney} provide users with easy access to tools that generate highly realistic and deceptive images. The misuse of these models poses significant concerns regarding the spread of misinformation, which can mislead the public and cause societal issues.  As a result, detecting images generated by diffusion models has become a critical challenge in maintaining trust in digital content.

Many existing methods \cite{luo2021generalizing, jeong2022bihpf, jeong2022frepgan, ju2023glff, deng2023new, tan2024frequency} have shown promising results in detecting images generated by the same model used for training. However, these approaches exhibit limited generalization when applied to images from unseen diffusion models. In such cases, performance often drops drastically, highlighting a critical challenge for current detection methods. Several existing methods \cite{tan2023learning, wang2023dire, luo2024lare} attempt to enhance the generalization of detectors by finding shared forgery features across images generated by different models. Although these approaches show some improvements in generalization, their performance remains limited. This raises a question:
\begin{center}
\textbf{Can we identify a universal feature that effectively distinguish images generated by different diffusion models from real images?}
\end{center}

Based on the above question, we performed an experimental analysis using the GenImage \cite{zhu2024genimage} dataset to explore the generalization capability of detectors. Specifically, we constructed the training dataset by randomly selecting 1,600 real and 1,600 generated images from each category. The detectors were then evaluated on a test dataset comprising images generated by eight different models. The results are presented in Figure \ref{img3}.

We compared the evaluation feature space of the standard ResNet50 trained on images generated by a single model (ADM) versus images generated by all involved models, as shown in Figure \ref{img3} (a) and (b). Detectors trained on images from a single model type can effectively identify images generated by the same model but struggle with detecting images from unseen models. In contrast, detectors trained on images from all involved model types can accurately distinguish between real and generated images across different models. These results confirm the existence of a universal feature that effectively distinguishes images generated by various diffusion models from real images. However, a detector trained on images from a single model type struggles to learn this universal feature, limiting its generalization across different generative models. This limitation arises because the detector tends to overfit specific forgery patterns \cite{corvi2023intriguing} unique to its training set, which are not present in images generated by other diffusion models. 

Therefore, to extract the universal distinctions between images generated by different diffusion models and real images, we considered a pre-trained model trained on a large-scale dataset. As the model is trained on a large number of real images, it learns rich feature representations of them. Consequently, when extracting image features, the model can effectively cluster those of real images, as shown in Figure \ref{img3} (c). Therefore, we believe that, although different diffusion models may exhibit distinct patterns, all of them differ from real images. We can leverage the inherent strengths of pre-trained models to capture these differences and identify a universal feature that distinguishes real images from those generated by various models.

Based on the findings above, in this paper, we propose a effective training approach based on pre-trained models, termed Learning on Less (LoL). This method achieves strong generalization in detecting images generated by unseen various diffusion models (Figure \ref{img3} (d)) by: (1) Leveraging the pre-trained model's inherent ability to cluster real images, derived from its training on large-scale real-world datasets. (2) constraining the model's learning during training process to prevent it from acquiring patterns unique to a certain type of diffusion model. This enables the detector to focus on more generalized forgery features, enhancing its ability to detect diverse diffusion-generated images. 

To evaluate the effectiveness of our proposed method, we conducted extensive experiments on the GenImage dataset \cite{zhu2024genimage}. The results show that our method achieves state-of-the-art generalization performance using only a minimal amount of training data, significantly outperforming existing approaches.

Our main contributions are three-fold as follows:
\begin{itemize}[leftmargin=0.5cm, itemindent=0.3cm]
	\item[$\bullet$] We propose a simple yet effective training approach, Learning on Less (LoL), that improves the generalization of diffusion model-generated image detection by constraining the learning process of pre-trained models.
	\item[$\bullet$] Drawing inspiration from the concept of attention masking in NLP, we propose a mask generation algorithm and analyze how zero-masking effectively prevents the pre-trained model from overfitting to the training data.
	\item[$\bullet$] Extensive experiments demonstrate that even with just \textbf{1\%} of the training data, our method outperforms the current state-of-the-art \cite{luo2024lare}, achieving up to a \textbf{13.6\%} improvement in Average ACC under optimal conditions.
\end{itemize}

\section{Related Work}
\label{sec:relatedwork}
With the rapid advancement of image generation technology, detection methods for generated images have also significantly progressed in recent years to meet this emerging challenge. Early studies \cite{agarwal2017photo, mccloskey2018detecting, mccloskey2019detecting, nataraj2019detecting, li2020identification} used traditional detection methods based on handcrafted features, leveraging artifacts like color discrepancies, compression patterns, and saturation cues in generated images. As deep learning advanced, researchers found CNN-based methods achieved outstanding performance, shifting focus toward deep learning-based approaches. Previous methods based on spatial \cite{liu2020global, wang2020cnn, ju2023glff, deng2023new} and frequency \cite{frank2020leveraging, luo2021generalizing, jeong2022bihpf, jeong2022frepgan, tan2024frequency} domains demonstrated high accuracy in detecting images from the same generative models but struggled to identify images generated by unseen models \cite{cozzolino2018forensictransfer, zhang2019detecting}. 

To address the issue of model generalization, recent methods detect generated images by identifying forged features universally applicable across different generative models. For instance, Liu \textit{et~al.} \cite{liu2022detecting} have discovered that noise patterns in real images exhibit similar characteristics in the frequency domain, unlike generated images, and proposed a generalized feature, called Learned Noise Patterns. LGrad \cite{tan2023learning} transforms an RGB image into its gradient, serving as a general feature representation. Tan \textit{et~al.} \cite{tan2024rethinking} introduced the neighboring pixel relationship (NPR) to capture artifacts left by the upsampling process during image generation. Furthermore, FatFormer \cite{liu2024forgery} incorporates a contrastive learning objective between image features and text embeddings to enhance generalization capabilities. Ojha \textit{et~al.} \cite{ojha2023towards} and Koutlis \textit{et~al.} \cite{koutlis2024leveraging} freeze the pre-trained CLIP-ViT \cite{radford2021learning} encoder, feeding the extracted features into a classification head for generated image detection. This approach mitigates the risk of the model overfitting to the training data. SeDIE \cite{ma2023exposing}, DIRE \cite{wang2023dire}, and $\text{LaRE}^{2}$ \cite{luo2024lare} leverage reconstruction errors from diffusion models to enable generalized detection of diffusion-generated images. In contrast to these works, we rethink the effectiveness of pre-trained models in enhancing the generalization of diffusion model detection.  By leveraging pre-trained weights trained on large-scale, real-world images to help detectors achieve better generalization.

\section{Methodology}
\label{sec:method}

\subsection{Problem Definition}
To address the current generalization challenges, we aim to develop a generalizable detector capable of accurately identifying images generated by various diffusion models. In this context, we define $I = \{I_{1}, I_{2}, \cdots, I_{n}\}$ as the set of images generated by $n$ different diffusion models. Each image $I_{k}^{i} \in I_{k}$ is associated with a label $y_{k}^{i} \in \{0,1\}$, where $y_{k}^{i} = 1$ indicates the real image and $y_{k}^{i} = 0$ indicates the generated image. Assuming an optimal generalization solution $\theta^{*}$ exists for the model parameters, it is able to identify images generated by different diffusion models, represented as:

\begin{equation}\label{eqn-1} 
	\begin{aligned}
		\exists \theta^* \text{ such that } & \min_{\theta} Loss(D(I_k; \theta), y_k), \\
		 \forall I_k \in I, \; &k = 1, 2, \ldots, n
	\end{aligned}
\end{equation}
where $\min_{\theta} Loss$ represents the minimum loss of detecting images generated by different models, $D$ denotes the classifier, $\theta$ represents the model parameters after training.

Our goal is to ensure that the detector maintains high detection accuracy, even for images generated by previously unseen diffusion models. Specifically, for training images $I_{k}$ generated by a given diffusion model $k$, we aim to find a set of model parameters $\theta_{I_k}$ that is as close as possible to the optimal solution $\theta^*$, which can be expressed as:

\begin{equation}\label{eqn-2} 
	\min_{\theta} ||\theta_{I_k} - \theta^*||
\end{equation}

\subsection{Analysis Behind the Method}
Images generated by different diffusion models exhibit unique artifacts \cite{corvi2023intriguing}, making it challenging for a model trained on one type of image to accurately detect images generated by unseen models. In order to approach the optimal solution $\theta^{*}$ for generalization, we consider levering the pre-trained models, trained on large-scale datasets, to extract universal forgery traces. 

\begin{figure}[t]
	\centering
	\includegraphics[width=8.2cm]{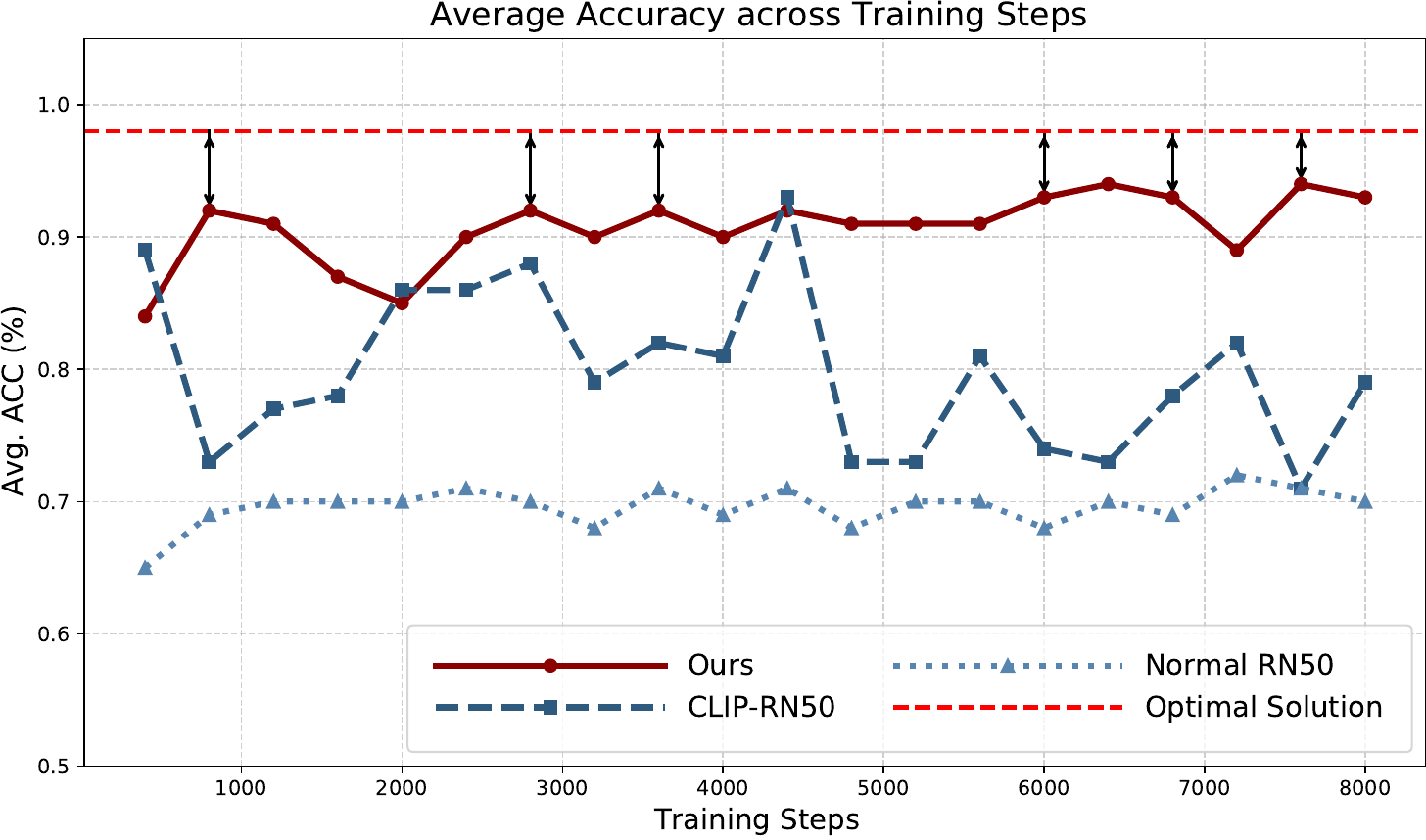}
	\caption{Average Accuracy Across GenImage Test Sets for Models Trained on ADM. Performance comparison at various training steps for Normal ResNet50, pre-trained CLIP-ResNet50, and pre-trained CLIP-ResNet50 using our proposed method.}
	\label{img7}
\end{figure}

\subsubsection{Generalization of Pre-trained Models}
Given that CLIP \cite{radford2021learning} is trained on 400 million image-text pairs, we utilize its pre-trained image encoder to extract generalized features, which are then passed through a classification head for final classification. To assess the effectiveness of pre-trained models, we tested both a standard ResNet50 and pre-trained CLIP-RN50 as classifiers. The models were trained on the ADM dataset from GenImage \cite{zhu2024genimage}. To monitor the models' performance, we conducted evaluations on the GenImage test set, which includes images generated by eight different models, recording results in every 400 training steps. The results of these evaluations are presented in Figure \ref{img7}. 

The experimental results demonstrate that: (1) Compared with the retrained ResNet50, the ResNet50 based on model weights pre-trained on a large-scale dataset exhibits significantly enhanced generalization ability. Specifically, the pre-trained CLIP-RN50, benefiting from extensive pre-training, approaches the optimal solution at certain points in training, confirming that utilizing a pre-trained model can indeed guide the model parameters $\theta$ toward the optimal solution $\theta^{*}$ for generalization. (2) However, the performance of the pre-trained CLIP-RN50 exhibits some instability, which can be attributed to distinct patterns in the training dataset that occasionally cause the model to deviate from the optimal solution.

\begin{figure*}[t]
	\centering
	\includegraphics[width=17.5cm]{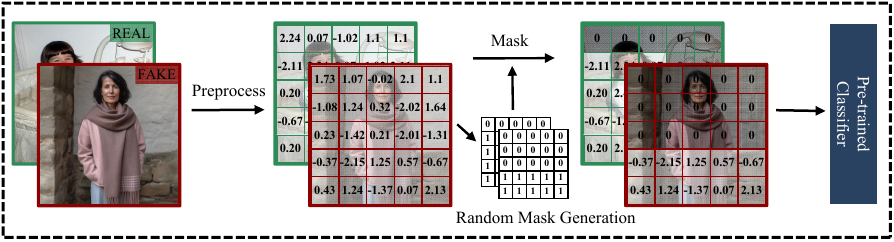}
	\caption{Overview of the Constrained Learning Process Pipeline.}
	\label{img8}
\end{figure*}

\subsubsection{Effect of Masking on Learning Constraints}
To address model performance fluctuations, we drew inspiration from the attention mask mechanism \cite{vaswani2017attention} in the field of natural language processing (NLP). The masked attention mechanism governs which parts of the input the model attends to and which it ignores during training. This approach enables the model to focus on specific areas, thereby shielding part of information. The mask is defined as follows:

\begin{equation}\label{eqn-3} 
M_{ij} =
\begin{cases}
	0,  & \text{Allowing $j$ to attend to $i$} \\
	-\infty, & \text{Prohibiting $j$ from attending to $i$}
\end{cases}
\end{equation}

Introducing it into the attention score to achieve the function of shielding information, as shown below:

\begin{equation}\label{eqn-4} 
\text{Attention}(Q, K, V) = \text{softmax}\left(\frac{QK^T}{\sqrt{d_k}} + M\right)V
\end{equation}

In NLP, negative infinity values are commonly used to disregard specific areas within the attention mechanism In this work, we adapt this concept by setting certain pixel values to zero, allowing the model to ignore specific regions. To illustrate the effectiveness of this operation, we present a comprehensive analysis in this section.

Consider the backpropagation process in convolutional neural networks (CNNs) \cite{rumelhart1986learning, zhang2016derivation}, given the prediction of the model $\hat{y}$ and ground-truth label $y$, we calculate loss using the standard Binary Cross Entropy (BCE), which is represented as:

\begin{equation}\label{eqn-5} 
	L = - \left( y \log(\hat{y}) + (1 - y) \log(1 - \hat{y}) \right)
\end{equation}

Assuming the input image is denoted as $X=\{x_{ij}\}$ the output from the first convolutional layer is $Z^{1}=\{z^{1}_{rc}\}$, we focus on this initial layer during the gradient calculation in back-propagation. To ensure clarity in our analysis, we simplify the process by considering only a single channel of the image, as the same operation applies to each channel. The gradient for each parameter in this layer is computed as follows:

\begin{equation}\label{eqn-6} 
	\begin{aligned}
		z^{1}_{rc} &= \sum w^{1}_{mn}  \cdot x_{ij} + b \\
		\frac{\partial L}{\partial w^{1}_{mn}} &= \sum \frac{\partial L}{\partial z_{rc}} \cdot \frac{\partial z^{1}_{rc}}{\partial w^{1}_{mn}} = \sum x_{ij}
	\end{aligned}
\end{equation}
where $w^{1}_{mn}$ denoted the parameters, $x_{ij}$ and $z^{1}_{rc}$ represent the pixel value and the first convolution layer output related to each parameter, respectively.

Subsequently, the model parameters are updated by calculating the gradients of the parameters, as follows:
\begin{equation}\label{eqn-7} 
	{w^{1}_{mn}}^{*} = w^{1}_{mn} - \alpha \cdot \frac{\partial L}{\partial w^{1}_{mn}}
\end{equation}
where ${w^{1}_{mn}}^{*}$ denotes updated parameters, $\alpha$ represents the learning rate.

When an input pixel $x_{ij} = 0$, it does not contribute to the parameter update of the first convolutional layer. This implies that the first layer can directly sense and effectively ignore 0-value pixels. However, as the receptive field expands in subsequent layers, zero-value pixels may be influenced by surrounding non-zero pixels, resulting in their involvement in parameter updates. Despite this, their contribution remains significantly constrained. Consequently, setting specific regions of the input image to zero during training can substantially constrain the model's attention to these areas, thereby preventing it from learning excessive details from the input data.

\subsection{The Proposed Learning on Less Framework}
Building on the analysis above, we propose a simple yet effective training approach based on pre-trained models, called Learning on Less (LoL), which generates random masks during training and applies them to input tensors, as shown in Figure \ref{img8}. These masks selectively block out parts of the image, preventing the model from overfitting to specific details in the original training images and encouraging it to focus on more general features instead.

\subsubsection{Random Mask Generation}
\label{sec: mask_gen}
In order to force the model to ignore parts of the input image, we employ a masking mechanism similar to the attention mask used in NLP. Given an RGB image $X \in \mathbb{R}^{H\times W\times 3}$, we first generate a mask $M \in \mathbb{R}^{H\times W\times 1}$ with the same size as the input image, where each element is initially set to $M_{i,j} = 1 \quad \forall \, i \in [0, H-1], \, j \in [0, W-1]$. 
Next, we randomly select a region of the image $x$ corresponding to a specified proportion $S_{select} = H \times W \times r_{mask}$, where $r_{mask}$ represents the fraction of the image that will be ignored by the model. Using this region, we compute the dimensions of the ignored area, $H_{select}$ and $W_{select}$, based on the given aspect ratio $r_{aspect}$.

\begin{algorithm}
	\caption{Random Mask Generation}
	\label{algorithm1}
	\KwIn{An RGB image $x$ of size $H \times W \times 3$, Masked ratio $r_{mask}$, Aspect ratio $r_{aspect}$}
	\KwOut{A mask $M$ of size $H \times W \times 1$}
	
	\BlankLine
	\emph{Initialize mask $M$ with the size $H \times W \times 1$} \;
	\For{$i \gets 0$ \KwTo $H$}{
		\For{$j \gets 0$ \KwTo $W$}{
			Set $M[i, j] \gets 1$ \;
		}
	}
	
	$S_{select} \gets H \times W \times r_{mask}$ \;
	$H_{select} \gets round(\min(H, \sqrt{S_{select} \times r_{aspect}}))$ \;
	$W_{select} \gets round(\min(W, \sqrt{S_{select} / H_{select}}))$ \;
	
	$ t \gets randint(0, H - H_{select})$ \;
	$ l \gets randint(0, W - W_{select})$ \;
	
	\For{$i \gets t$ \KwTo $t + H_{select} - 1$} {
		\For{$j \gets l$ \KwTo $l + W_{select} - 1$} {
			Set $M[i, j] \gets 0$ \;
		}
	}
	\textbf{return} $M$;
\end{algorithm}

A corresponding region in the mask M is randomly selected, and the values within this region are set to zero, indicating that these areas of the image should be ignored. The coordinates of the upper-left corner of the selected region are determined through random generation. The pseudo-code for this mask generation process is provided in Algorithm \ref{algorithm1}, the output $M$ represents the binary mask corresponding to the input image, where $M_{ij} = 1$ indicates included pixels, and $M_{ij} = 0$ represents masked pixels.

The mask is then applied to instruct the model to ignore certain areas of the image, thereby enhancing generalization by preventing the model from overfitting to specific details.

\subsubsection{Constrained Learning Process Toward Optimal Generalization}
During the training process, the input image undergoes preprocessing steps, including cropping, tensor conversion, and normalization, which transform it into a suitable tensor format for model input. Before feeding the image into the model, a corresponding mask is generated and applied to its three channels to indicate regions of the image that should be ignored, as shown in:

\begin{equation}\label{eqn-8} 
	\begin{aligned}
		&X_{tensor} = \text{Preprocess}(X) \\
		&M = \text{Random Mask Generation}(X_{tensor}) \\
		&X_{input} = M \circ X_{tensor}
	\end{aligned}
\end{equation}
In this equation, $X_{tensor}$ represents the preprocessed image tensor, and $M$, where $M_{ij} \in \{0, 1\}$,  is the binary mask generated by the method described in Section \ref{sec: mask_gen}. $X_{input}$ denotes the final input tensor fed into the model.

\begin{figure*}[b]
	\centering
	\includegraphics[width=17.5cm]{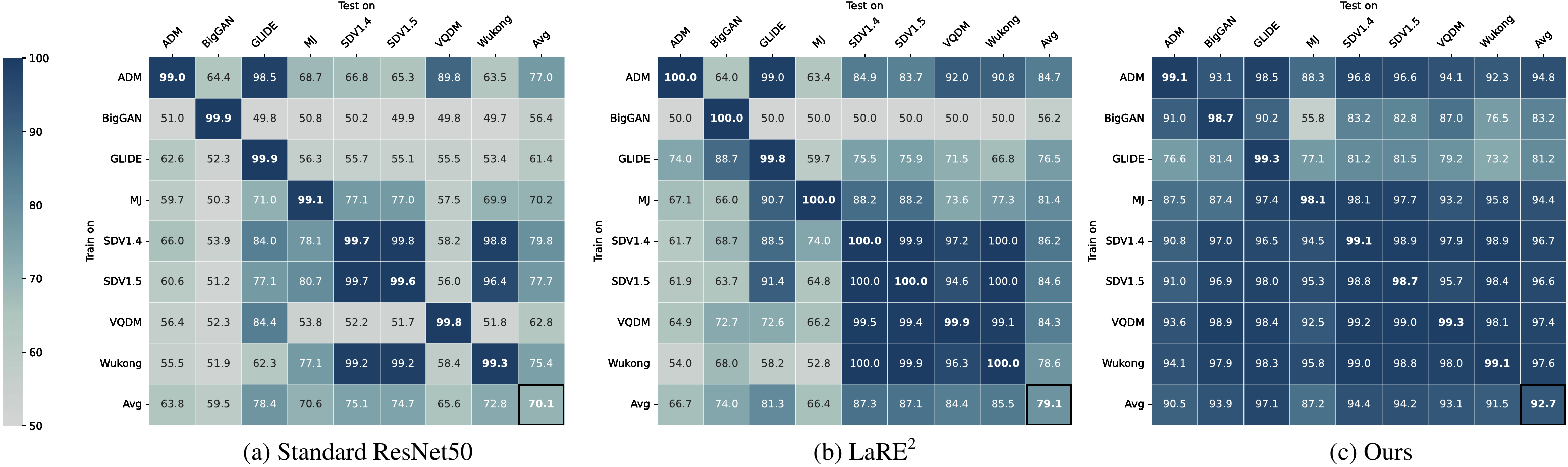}
	\caption{Accuracy (ACC) Results Across 8 Subsets. Each model is trained on a single subset and evaluated across all 8 subsets. The comparison includes the standard ResNet50 \cite{he2016deep}, $\text{LaRE}^{2}$ \cite{luo2024lare}, and our proposed method. The color scale reflects performance,  with darker shades representing higher accuracy values.}
	\label{img4}
\end{figure*} 

\begin{figure*}[t]
	\centering
	\includegraphics[width=17.5cm]{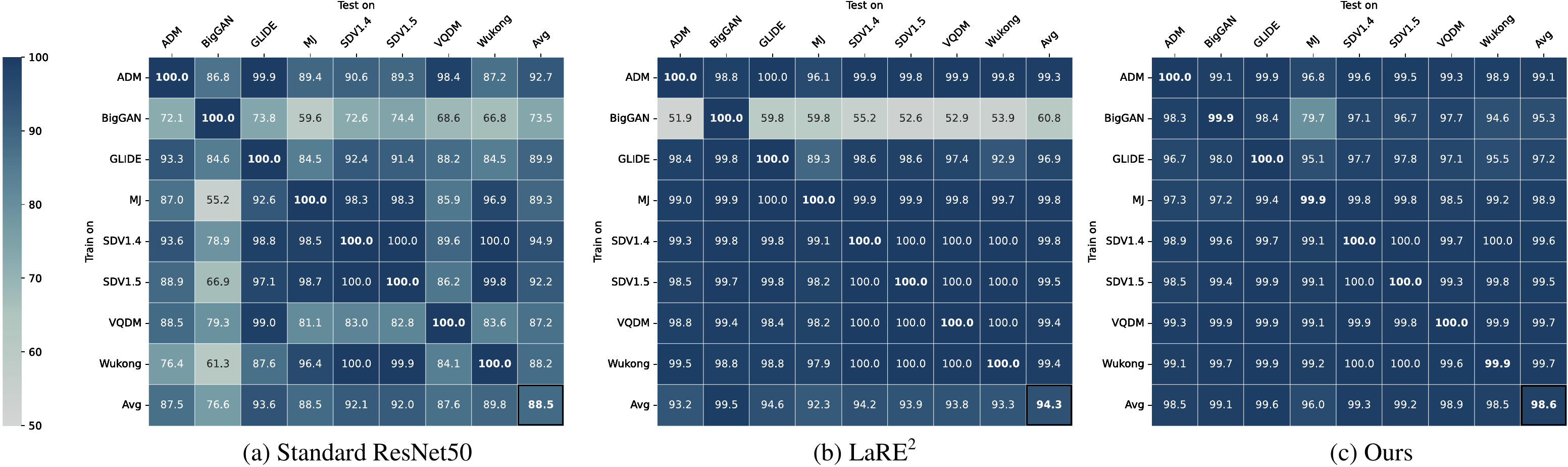}
	\caption{Average Precision (AP) Results Across 8 Subsets. Evaluation of the standard ResNet50, $\text{LaRE}^{2}$, and our proposed method across all 8 subsets, performed using the same approach as the accuracy (ACC) assessment. }
	\label{img5}
\end{figure*} 

This process effectively mitigates the influence of unique patterns specific to a certain type of diffusion model in the training set by constraining the model's learning from excessive details, allowing the model's parameters $\theta$ to smoothly approach the optimal solution $\theta^{*}$ throughout the training.

\section{Experiments}

\label{sec:experiments}
\subsection{Dataset}
To simulate real-world scenarios involving images of varying resolutions, we use the GenImage dataset \cite{zhu2024genimage} to evaluate our approach. The GenImage dataset consists of 1,331,167 real images and 1,350,000 fake images across various resolutions, organized into eight subsets. Each subset contains fake images generated by one of eight distinct generative models: AMD \cite{dhariwal2021diffusion}, BigGAN \cite{brock2018large}, GLIDE \cite{nichol2021glide}, Midjourney \cite{midjourney}, Stable Diffusion V1.4 \cite{rombach2022high}, Stable Diffusion V1.5 \cite{rombach2022high}, VQDM \cite{gu2022vector}, and Wukong \cite{wukong}. In our experiments, we follow the official dataset split, training on one subset of generated images and evaluating the generalization performance across all eight subsets. To further assess the impact of training data size on the generalization capabilities of pre-trained models, we randomly sample varying proportions of images from each generated dataset in the training set for our training configuration.

\begin{table*}[t]
	\centering
	\renewcommand{\arraystretch}{1.2}
	\setlength{\tabcolsep}{2.0mm}{
		\begin{tabular}{c|c c c c c c c c|c}
			\Xhline{1.4pt}
			
			\multirow{2}{*}{Methods} & \multicolumn{8}{c|}{Testing Subset} & \multirow{2}{*}{Avg. Acc.(\%)}\\
			\cline{2-9}
			\multicolumn{1}{c|}{} & ADM & BigGAN & GLIDE & MidJourney & SDV1.4 & SDV1.5 & VQDM & Wukong & \multicolumn{1}{c}{} \\
			\cline{1-10}
			\Xhline{1.4pt}
			CNNSpot \cite{wang2020cnn} & 57.0 & 56.6 & 57.1 & 58.2 & 70.3 & 70.2 & 56.7 & 67.7 & 61.7 \\
			\Xhline{0.8pt}
			Spec \cite{zhang2019detecting} & 57.9 & 64.3 & 65.4 & 56.7 & 72.4 & 72.3 & 61.7 & 70.3 & 65.1 \\
			\Xhline{0.8pt}
			F3Net \cite{qian2020thinking} & 66.5 & 56.5 & 57.8 & 55.1 & 73.1 & 73.1 & 62.1 & 72.3 & 64.6 \\
			\Xhline{0.8pt}
			GramNet \cite{liu2020global}  & 58.7 & 61.2 & 65.3 & 58.1 & 72.8 & 72.7 & 57.8 & 71.3 & 64.7 \\
			\Xhline{0.8pt}
			DIRE \cite{wang2023dire} & 61.9 & 56.7 & 69.1 & 65.0 & 73.7 & 73.7 & 63.4 & 74.3 & 67.2 \\
			\Xhline{0.8pt}
			$\text{LaRE}^{2}$ \cite{luo2024lare} & 66.7 & 74.0 & 81.3 & 66.4 & 87.3 & 87.1 & 84.4 & 85.5 & 79.1 \\
			\Xhline{1.4pt}
			\rowcolor[rgb]{0.96078, 0.96078, 0.96078}
			Ours  & \textbf{90.5} & \textbf{93.9} & \textbf{97.1} & \textbf{87.2} & \textbf{94.4} & \textbf{94.2} & \textbf{93.1} & \textbf{91.5} & \textbf{92.7} \\
			\Xhline{1.4pt}
		\end{tabular}
	}
	\caption{Comparison of Average Accuracy (Avg. ACC) between Our Method and Other Generated Image Detectors on the GenImage Test Set. Each model is trained using data from eight generators and evaluated across all test sets. Accuracy is averaged over the eight training cases per test set, with top-performing results highlighted in bold.}
	\label{table1}
\end{table*}

\subsection{Implementation Details}
We employed CLIP-RN50\cite{radford2021learning} as a classifier in our experiments. For training, input images were randomly cropped to 224 × 224, with horizontal flipping and rotation applied for data augmentation. Images that did not meet the minimum crop requirements were expanded by stitching repeated content to achieve the necessary crop size. In contrast, only center cropping was applied during testing. The Adam optimizer \cite{kingma2014adam} with beta parameters (0.9, 0.999) was employed to minimize binary cross-entropy loss. The models were trained with a learning rate of \num{5e-6} for 20 epochs and a batch size of 4. For the BigGAN training process, a lower learning rate of \num{5e-7} was used to ensure stability. All experiments were conducted using the PyTorch framework \cite{paszke2019pytorch} on an Nvidia GeForce RTX 3090 GPU.

\subsection{Evaluation metric}
In accordance with the protocols outlined in DIRE and $\text{LaRE}^{2}$, we use Accuracy (ACC) and Average Precision (AP) as the primary evaluation metrics. To assess the model's generalization capability, we train it on one subset and evaluate it across all eight subsets, computing the average ACC and AP. These metrics provide a comprehensive characterization of the model's generalization performance across diverse datasets.

\begin{figure}[htb]
	\centering
	\includegraphics[width=8.2cm]{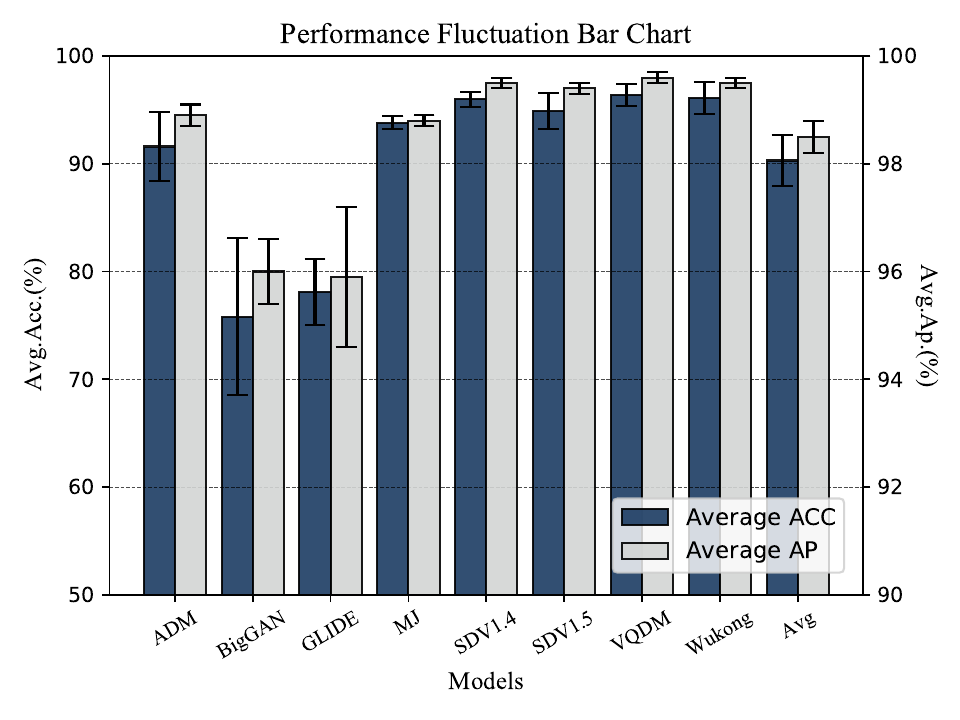}
	\caption{Performance Fluctuation Bar Chart. Average Acc and Ap across different training cases tested on all GenImage test sets, with error bars included.}
	\label{img6}
\end{figure}

\subsection{Generalization Evaluation}
To evaluate the generalization capability of our proposed method, we conducted experiments on the GenImage dataset using both the standard ResNet50 and our approach. The models were trained on one subset and evaluated across all eight subsets.

As shown in Figures \ref{img4} and \ref{img5}, our method exhibits exceptional generalization performance. Training on each subset with only 1,600 real and 1,600 generated images, we achieved an average accuracy (AvgAcc) of 92.7\% and an average precision (AP) of 98.6\%. In comparison, our method outperformed the standard ResNet50 by a substantial margin of 22.7\% in AvgAcc.
Notably, even when trained on BigGAN-generated images, our method achieved an AvgAcc of 83.2\% across all remaining subsets, which, except for the BigGAN subset, are based on diffusion models. This demonstrates the superior generalization capability of our approach across various generative models.

To further evaluate the effectiveness of our proposed method, we compared its performance against recent state-of-the-art methods DIRE\cite{wang2023dire} and $\text{LaRE}^{2}$ \cite{luo2024lare}, as well as several classic approaches discussed in $\text{LaRE}^{2}$, including CNNSpot \cite{wang2020cnn}, Spec \cite{zhang2019detecting}, F3Net \cite{qian2020thinking}, and GramNet \cite{liu2020global}.
As shown in Table \ref{table1}, our method outperforms all others across all eight subsets, significantly outperforming the state-of-the-art method $\text{LaRE}^{2}$, with a 13.6\% improvement in AvgAcc across all subsets. Even in the BigGAN-trained experiment set, where the $\text{LaRE}^{2}$'s performance drastically declined, our method sustained strong results, with an AvgAcc 27.0\% higher and an AvgAP 34.5\% higher, as illustrated in Figures \ref{img4} and \ref{img5}. These results demonstrate the effectiveness of our approach.

\begin{table*}[ht]
	\centering
	\renewcommand{\arraystretch}{1.2}
	\setlength{\tabcolsep}{2.7mm}{
		\begin{tabular}{c|c c|c c|c c}
			\Xhline{1.4pt}
			
			\multirow{3}{*}{\makecell{Data Volume \\ count (proportion \%)}} & \multicolumn{6}{c}{Model} \\
			\cline{2-7}
			\multicolumn{1}{c|}{} & \multicolumn{2}{c|}{CLIP-RN50} & \multicolumn{2}{c|}{CLIP-RN101} & \multicolumn{2}{c}{CLIP-ViT-L/14} \\
			\cline{2-7}
			\multicolumn{1}{c|}{} & Avg.Acc.(\%) & Avg.Ap.(\%) & Avg.Acc.(\%) & Avg.Ap.(\%) & Avg.Acc.(\%) & Avg.Ap.(\%) \\
			\cline{1-4}
			\Xhline{1.4pt}
			400 (0.125\%) & 92.6 ($\pm$1.6) & 98.5 ($\pm$0.2) & 90.8 ($\pm$1.1) & 97.6 ($\pm$0.1) & \cellcolor{lightgray}\textbf{95.4 ($\pm$1.5)} & \cellcolor{lightgray}\textbf{99.1 ($\pm$0.2)} \\
			\Xhline{0.8pt}
			800 (0.25\%) & \textbf{94.5 ($\pm$0.9)} & \textbf{98.8 ($\pm$0.2)} & 93.1 ($\pm$2.0) & 98.9 ($\pm$0.1) & 91.4 ($\pm$3.8) & 99.4 ($\pm$0.1) \\
			\Xhline{0.8pt}
			1,600 (0.5\%) & \textbf{94.5 ($\pm$0.8)} & \textbf{99.2 ($\pm$0.0)} & 93.2 ($\pm$1.2) & 98.5 ($\pm$0.1) & 94.7 ($\pm$2.4) & 99.7 ($\pm$0.1) \\
			\Xhline{0.8pt}
			3,200 (1.0\%) & \cellcolor{lightgray}\textbf{96.0 ($\pm$0.7)} &\cellcolor{lightgray} \textbf{99.5 ($\pm$0.1)} & 90.3 ($\pm$3.1) & 98.2 ($\pm$0.1) & 89.0 ($\pm$6.3) & 99.3 ($\pm$0.2) \\
			\Xhline{0.8pt}
			6,400 (2.0\%) & \textbf{95.7 ($\pm$0.9)} & \textbf{99.6 ($\pm$0.0)} & 92.5 ($\pm$2.9) & 99.3 ($\pm$0.1) & 87.0 ($\pm$10.2) & 97.9 ($\pm$1.7) \\
			\Xhline{0.8pt}
			12,800 (4.0\%) & 95.5 ($\pm$1.9) & 99.4 ($\pm$0.3) & \cellcolor{lightgray}\textbf{96.2 ($\pm$0.6)} & \cellcolor{lightgray}\textbf{99.5 ($\pm$0.1)} & 84.5 ($\pm$6.4) & 99.3 ($\pm$0.5) \\
			\Xhline{0.8pt}
			25,600 (8.0\%) & 94.0 ($\pm$2.2) & 99.4 ($\pm$0.1) & \textbf{94.0 ($\pm$1.5)} & \textbf{99.4 ($\pm$0.1)} & 87.1 ($\pm$9.0) & 97.6 ($\pm$2.1) \\
			\Xhline{0.8pt}
			64,000 (20.0\%) & 92.5 ($\pm$4.5) & 99.6 ($\pm$0.2) & \textbf{92.5 ($\pm$3.8)} & \textbf{99.4 ($\pm$0.1)} & 84.6 ($\pm$6.7) & 99.6 ($\pm$0.1) \\
			\Xhline{0.8pt}
			160,000 (50.0\%) & \textbf{93.4 ($\pm$1.5)} & \textbf{99.7 ($\pm$0.1)} & 93.6 ($\pm$2.9) & 99.6 ($\pm$0.1) & 85.7 ($\pm$4.9) & 99.6 ($\pm$0.1) \\
			\Xhline{0.8pt}
			323,994 (100.0\%) & \textbf{89.9 ($\pm$3.3)} & \textbf{99.7 ($\pm$0.0)} & 90.0 ($\pm$4.6) & 99.7 ($\pm$0.1) & 82.1 ($\pm$5.3) & 98.0 ($\pm$1.4) \\
			\Xhline{1.4pt}
		\end{tabular}
	}
	\caption{Average Acc and Ap Across GenImage Test Sets for Different Model Architectures (CLIP-RN50, CLIP-RN101, CLIP-ViT L/14) Trained with Varying Data Volumes. The best results for each model architecture are shaded in gray. The top-performing results for each data volume are highlighted in bold.}
	\label{table3}
\end{table*}

Additionally, to further assess the stability of the model's performance, we evaluated the weights from the final five epochs after the model had stabilized during training, as shown in Figure \ref{img6}. The experimental results demonstrate that our method exhibits notable stability, with the average accuracy (ACC) across all subsets fluctuating around 90\%. Even in the worst case, the model achieved an ACC of 87.9\%, which is 8.8\% higher than the current state-of-the-art method, $\text{LaRE}^2$. Notably, the model's performance was most stable when trained on the MidJourney and Stable Diffusion V1.4 datasets, whereas it exhibited greater fluctuation when trained on the BigGAN and GLIDE datasets. We attribute this to the quality of the generated images, as both BigGAN and GLIDE produce images of lower quality.

\subsection{Ablation Study}
In this section, we conduct comprehensive ablation studies to analyze the impact of various factors, including model architecture, training data volume, masked ratio, and aspect ratio, on generalization performance. All experiments are performed on the most stable training set SDV1.4, the same as mentioned in GenImage.

\subsubsection{Impact of Training Data Volume}
As shown in Figure \ref{img7}, our method achieves excellent results after only 3,000 training steps, suggesting that excessive data may not be required under pretraining conditions. To further investigate the impact of training data volume, we analyzed the model's performance with varying amounts of data, as shown in Table \ref{table3}. For stability assessment, we also evaluated the model's performance using weights from the five epochs after it reached training stability. The results indicate that, with CLIP-RN50 as the classifier, our method performs optimally with only 1\% of the training dataset. Interestingly, as the data volume increases, the model's performance declines and becomes unstable, likely due to overfitting to patterns unique to the training set.

\subsubsection{Influence of Model Architecture}
To assess the impact of different model architectures on generalization, we used CLIP-RN50, CLIP-RN101, and CLIP-ViT-L/14 as classifiers to evaluate the effectiveness of our method. For each classifier, we conducted experiments with varying data volumes, as shown in Table \ref{table3}. The results indicate that CNN-based architectures outperform Transformer-based architectures in terms of both performance and stability. Additionally, as the model size increases, its data requirements to achieve optimal performance also increase. For example, the larger CLIP-RN101 model necessitates more data to achieve optimal results.

\begin{table}[t]
	\centering
	\renewcommand{\arraystretch}{1.2}
	\setlength{\tabcolsep}{1.5mm}{
		\begin{tabular}{c|c c c}
			\Xhline{1.4pt}
			
			\multirow{2}{*}{\makecell{Masked Ratio \\ $r_{mask}$}} & \multicolumn{3}{c}{Aspect Ratio $r_{aspect}$} \\
			\cline{2-4}
			\multicolumn{1}{c|}{} & (1.0, 1.0) & (0.5, 2.0) & (0.33, 3.0) \\
			\cline{1-4}
			\Xhline{1.4pt}
			(0.0, 0.2) & 91.6 ($\pm$3.6) & 93.9 ($\pm$1.4) & 94.0 ($\pm$1.0) \\
			\Xhline{1.4pt}
			(0.2, 0.4) & 92.4 ($\pm$2.0) & 93.7 ($\pm$1.3) & 94.3 ($\pm$1.0) \\
			\Xhline{1.4pt}
			(0.4, 0.6) & 94.5 ($\pm$2.1) & 95.3 ($\pm$1.6) & 94.5 ($\pm$1.5) \\
			\Xhline{1.4pt}
			\cellcolor{lightgray}(0.6, 0.8) & \cellcolor{lightgray}94.4 ($\pm$1.5) & \cellcolor{lightgray}95.9 ($\pm$0.9) & \cellcolor{lightgray}\textbf{96.0 ($\pm$0.7)} \\
			\Xhline{1.4pt}
			(0.8, 1.0) & 92.4 ($\pm$0.9) & 95.7 ($\pm$0.4) & 95.8 ($\pm$1.1) \\
			\Xhline{1.4pt}
		\end{tabular}
	}
	\caption{Average Acc across different training cases tested on all GenImage test sets, with variations in masked ratio and aspect ratio.}
	\label{table4}
\end{table}
\subsubsection{Effect of Masked Ratio and Aspect Ratio}
In this section, we conducted ablation experiments to examine the effects of the Mask Ratio $r_{mask}$ and Aspect Ratio $r_{aspect}$ within the Random Mask Generation algorithm. To increase data variability during training, each ratio was randomly selected within a specified range. The results, presented in Table \ref{table4}, demonstrate that greater diversity in aspect ratios leads to more stable and improved model performance. Furthermore, for the mask ratio, covering 60\%-80\% of the image area achieved the best results. This range effectively enabled the model to learn generalizable features for distinguishing real and generated images while preventing overfitting to the training data.

\section{Conclusion}
\label{sec:Conclusion}

In this paper, we introduce a effective training approach, Learning on Less (LoL), which enhances the detection of diffusion-generated images by constraining the model's learning. Leveraging the inherent generalization capabilities of pre-trained weights, our method enables the model to converge steadily toward an optimal solution for generalization. Experimental results demonstrate that our approach achieves state-of-the-art performance, showcasing exceptional generalization ability with a remarkably small amount of training data. 

\noindent{\textbf{Limitations.}} Admittedly, our method is sensitive to the quality of the generated images used in training. Lower image quality can affect its performance. In future work, we aim to develop more robust methods capable of maintaining strong performance, even with lower-quality training images.

{
    \small
    \bibliographystyle{ieeenat_fullname}
    \bibliography{main}
}

\clearpage
\setcounter{page}{1}
\maketitlesupplementary
\appendix
\renewcommand{\thetable}{A\arabic{table}}
\renewcommand{\thefigure}{A\arabic{figure}}

\section{Robustness Analysis}
In this section, we present robustness experiments to evaluate the effects of various perturbations on model performance, following the methodology outlined by Frank \textit{et~al.} \cite{frank2020leveraging}. These experiments are conducted on the stable SDV1.4 training set and evaluated on the GenImage \cite{zhu2024genimage} test sets to ensure consistent evaluation conditions.
\subsection{Perburbations}
\noindent{\textbf{Noise:}} Random Gaussian noise is added to the input images by selecting a variance value from a uniform distribution [5.0, 20.0], which controls the noise intensity. The result is a noisy image with the same dimensions as the original but with random variations in pixel intensity.

\noindent{\textbf{Blurring:}} Gaussian blur is applied to the input images with a randomly selected kernel size from the set [3,5,7,9]. Larger kernel sizes produce stronger blurring effects.

\noindent{\textbf{Compression:}} JPEG compression is applied to the input images by first selecting a random quality factor between 10 and 75. The image is then encoded into JPEG format using this quality factor, introducing lossy compression.

\noindent{\textbf{Cropping:}} The Random crop is applied to the input images by selecting a percentage between 5\% and 20\%, determining the crop size in both the x and y directions. The image is then resized to its original dimensions using cubic interpolation.

\begin{table*}[hbp]
	\centering
	\small
	\renewcommand{\arraystretch}{1.15}
	\setlength{\tabcolsep}{1.1mm}{
		\begin{tabular}{c|c c|c c|c c|c c}
			\Xhline{1.4pt}
			\multirow{3}{*}{\makecell{Data Volume \\ count (proportion \%)}} & \multicolumn{8}{c}{Perturbation} \\
			\cline{2-9}
			\multicolumn{1}{c|}{} & \multicolumn{2}{c|}{Noise} & \multicolumn{2}{c|}{Blurring} & \multicolumn{2}{c|}{Compression} & \multicolumn{2}{c}{Cropping} \\
			\cline{2-9}
			\multicolumn{1}{c|}{} & Avg.Acc.(\%) & Avg.Ap.(\%) & Avg.Acc.(\%) & Avg.Ap.(\%) & Avg.Acc.(\%) & Avg.Ap.(\%) & Avg.Acc.(\%) & Avg.Ap.(\%)\\
			\cline{1-4}
			\Xhline{1.4pt}
			1,600 (0.5\%) & 82.4 ($\pm$0.6) & 91.5 ($\pm$0.4) & 76.4 ($\pm$0.6) & 86.8 ($\pm$0.1) & 87.6 ($\pm$0.3) & 95.2 ($\pm$0.1) & 94.7 ($\pm$0.6) & 98.9 ($\pm$0.1)\\
			\Xhline{0.8pt} 
			3,200 (1.0\%) & 84.7 ($\pm$0.9) & 94.0 ($\pm$0.3) & 76.9 ($\pm$0.2) & 88.4 ($\pm$0.2) & 88.9 ($\pm$0.6) & 96.1 ($\pm$0.2) & 95.1 ($\pm$0.7) & 99.3 ($\pm$0.2)\\
			\Xhline{0.8pt}
			6,400 (2.0\%) & 84.1 ($\pm$0.7) & 93.1 ($\pm$0.9) & 77.8 ($\pm$0.5) & 89.3 ($\pm$0.3) & 88.8 ($\pm$0.9) & 96.5 ($\pm$0.2) & 95.4 ($\pm$1.0) & 99.4 ($\pm$0.1)\\
			\Xhline{0.8pt}
			12,800 (4.0\%) & 85.4 ($\pm$1.2) & 95.0 ($\pm$0.4) & 79.7 ($\pm$0.5) & 89.4 ($\pm$0.4) & 90.1 ($\pm$1.1) & 97.0 ($\pm$0.2) & 94.2 ($\pm$2.0) & 99.2 ($\pm$0.2)\\
			\Xhline{0.8pt}
			25,600 (8.0\%) & 84.3 ($\pm$0.6) & 93.6 ($\pm$0.1) & 79.2 ($\pm$1.0) & 91.4 ($\pm$0.2) & 89.6 ($\pm$1.9) & 96.9 ($\pm$0.5) & 93.7 ($\pm$2.0) & 99.1 ($\pm$0.2)\\
			\Xhline{1.4pt}
		\end{tabular}
	}
	\caption{Average Accuracy (Avg.Acc) and Average Precision (Avg.Ap) across GenImage test sets for CLIP-RN50, trained with varying data volumes and under different perturbations. }
	\label{tablea1}
\end{table*}

\subsection{Experimental Analysis}
To evaluate the robustness of our method under different perturbations, we applied the mentioned noise, blurring, JPEG compression, and random cropping independently on the training and test sets. Additionally, experiments were conducted with varying amounts of training data to comprehensively evaluate model robustness. The results are shown in Table \ref{tablea1}. The results show that the model’s performance is significantly affected by noise, blurring, and JPEG compression, which impact image quality. In particular, blurring has a substantial effect on performance, as it obscures most of the key information of the image. In contrast, random cropping has almost no impact on model performance, suggesting that our method is more sensitive to image quality while exhibiting strong robustness to geometric transformations.

\begin{figure}[t]
	\centering
	\includegraphics[width=8.2cm]{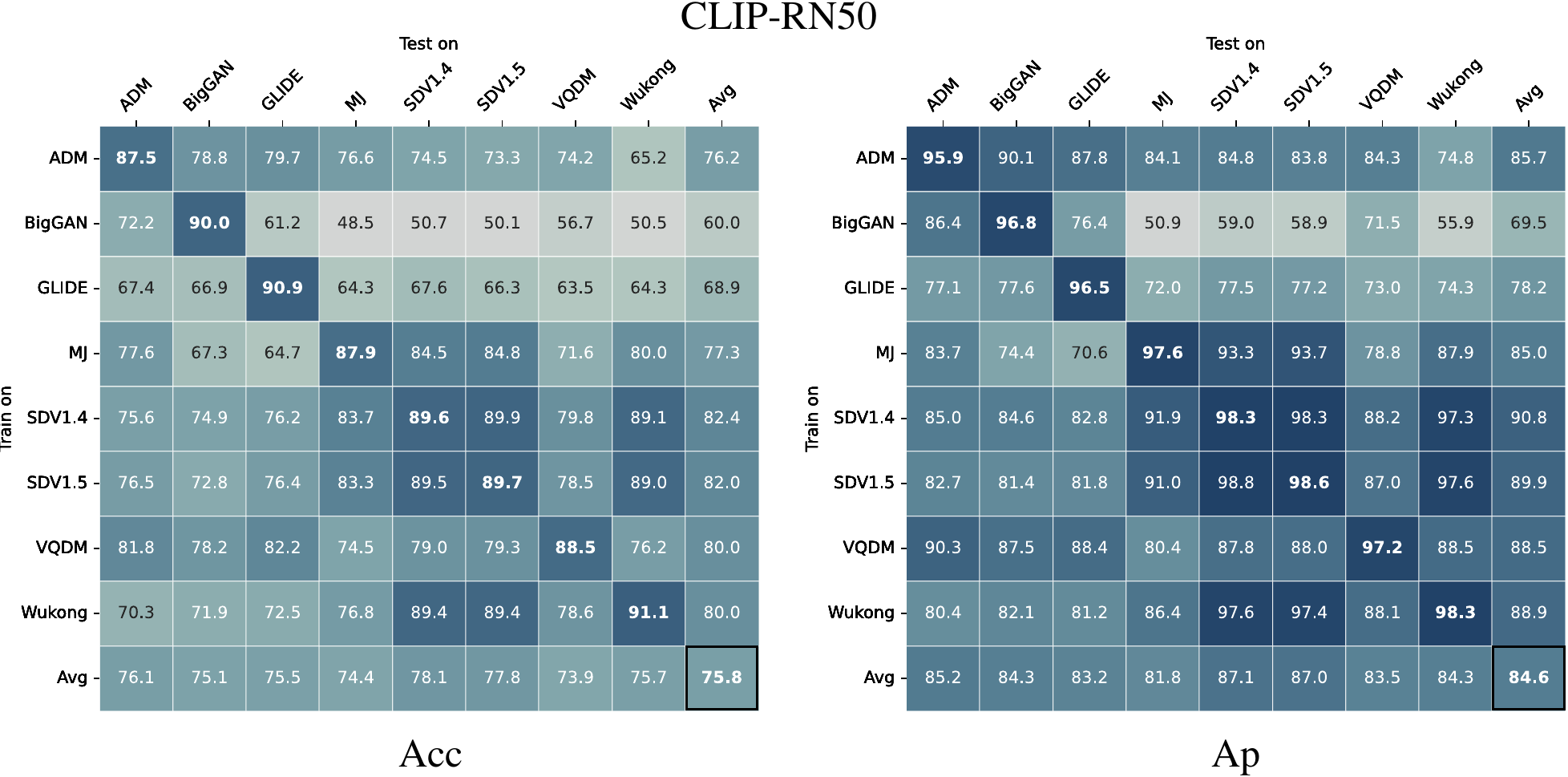}
	\caption{Accuracy (Acc) and Average Precision (Ap) results across eight subsets. The CLIP-RN50 model, trained on only 3,200 images, is evaluated across all eight subsets. The color scale represents performance, with darker shades indicating higher accuracy.}
	\label{imga1}
\end{figure}

\subsection{Evaluation in Real-world Scenarios}
To simulate conditions in real-world scenarios, we sequentially applied the four mentioned perturbations—noise, blurring, JPEG compression, and random cropping—on both the training and test sets. Additionally, we trained the model using only 1\% of the training data to simulate a data-scarce environment. The experimental results, shown in Figure \ref{imga1}, demonstrate that while the performance of our method degrades in this real-world scenario, it still maintains relatively high accuracy.

\begin{figure*}[t]
	\centering
	\includegraphics[width=17.5cm]{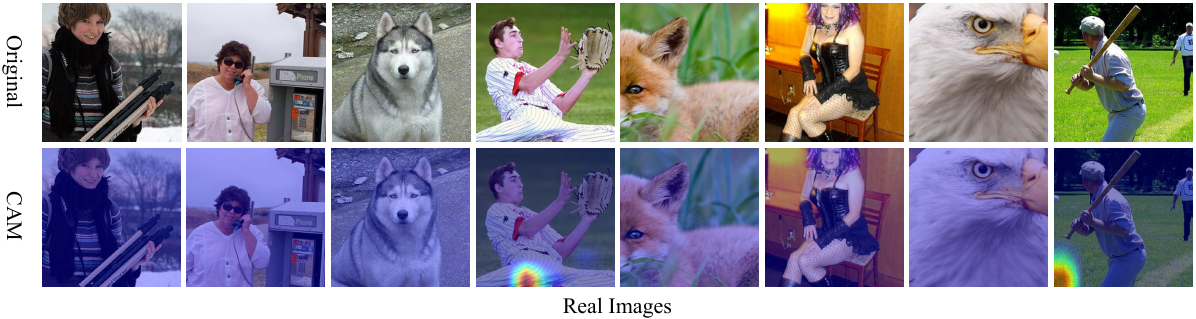}
	\caption{Class Activation Map (CAM) \cite{zhou2016learning} visualization extracted from the proposed LoL method on real images.}
	\label{imga2}
\end{figure*}

\begin{figure*}[t]
	\centering
	\includegraphics[width=17.5cm]{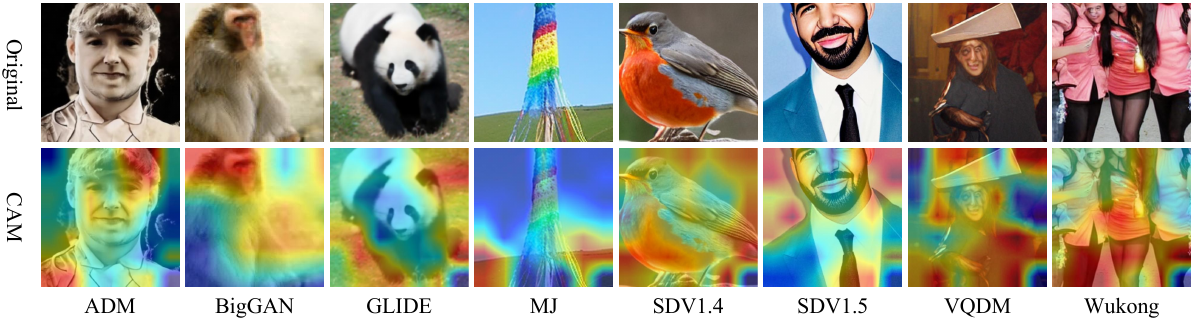}
	\caption{Class Activation Map (CAM) visualization extracted from the proposed LoL method on images generated by eight models in GenImage.}
	\label{imga3}
\end{figure*}

\section{Class Activation Map Visualization}
To further analyze how our proposed LoL method utilizes the pre-trained model to approach the optimal solution for generalization, we conducted class activation map (CAM) analysis on both real and generated images, as shown in Figure \ref{imga2} and \ref{imga3}. The visualization results reaffirm that there is a universal feature that can effectively distinguish real images from images generated by different models. Specifically, pre-trained models trained on large-scale real-world images can effectively cluster features of real images. As long as there are differences between generated and real images, even if the differences are not the same, the pre-trained models can capture those differences and achieve excellent generalization. Consequently, as shown in Figure \ref{imga2}, real images exhibit almost no response as their clusters exhibit no significant deviations. In contrast, images generated by different models exhibit a strong response, indicating that while each model produces unique patterns \cite{corvi2023intriguing}, all generated images differ fundamentally from real images.

{
}
\end{document}